\documentclass[12pt]{article}
\usepackage{amssymb,amsmath,latexsym}
\usepackage{soul,color}
\usepackage{xcolor}
\usepackage{graphicx,amssymb,amsmath}
\usepackage{algorithm}
\usepackage[noend]{algpseudocode}
\usepackage{pgfplots}
\usepackage{array}
\usepackage[utf8]{inputenc}
\newcolumntype{P}[1]{>{\centering\arraybackslash}p{#1}}
\usepackage{tikz}
\usepackage{caption}
\usepackage{pgfplots}
\usepackage{hyperref}
\usepackage{float}
\usepackage{subcaption}
\pgfplotsset{compat=newest}
\usepackage[utf8]{inputenc}
\usepackage{hyperref}
\usepackage{array}
\usepackage{graphicx}
\usepackage{multirow}
\usepackage{authblk}

\newcommand\MyBox[2]{
  \fbox{\lower0.75cm
    \vbox to 1.7cm{\vfil
      \hbox to 1.7cm{\hfil\parbox{1.4cm}{#1\\#2}\hfil}
      \vfil}%
  }%
}

\title{A Thousand Words are Worth More Than One Recording: NLP Based Speaker Change Point Detection}

\date{}

\author[1]{Or Haim Anidjar\thanks{Corresponding Author: orhaim.anidjar@g.ariel.ac.il}}
\author[2]{Chen Hajaj\thanks{chenha@g.ariel.ac.il}}
\author[3]{Amit Dvir\thanks{amitdv@g.ariel.ac.il}}
\author[2]{Issachar Gilad\thanks{giladis@g.ariel.ac.il}}
\affil[1]{Department of Computer Science, Ariel University, Israel.}
\affil[2]{Department of Industrial Engineering \& Management, Ariel University, Israel.}
\affil[3]{Ariel Cyber Innovation Center, Israel.}

\usepackage[T1]{fontenc}
\usepackage[utf8]{inputenc}
\usepackage{tabularx,ragged2e,booktabs,caption}
\newcolumntype{C}[1]{>{\Centering}m{#1}}

\begin{document}
	
	\maketitle
	
	\begin{abstract}
		Speaker Diarization (SD) consists of splitting or segmenting an input audio burst according to speaker identities. In this paper, we focus on the crucial task of the SD problem which is the audio segmenting process and suggest a solution for the Change Point Detection (CPD) problem. We empirically demonstrate the negative correlation between an increase in the number of speakers and the Recall and F1-Score measurements. This negative correlation is shown to be the outcome of a massive experimental evaluation process, which accounts its superiority to recently developed voice based solutions. In order to overcome the number of speakers issue, we suggest a robust solution based on a novel Natural Language Processing (NLP) technique, as well as a metadata features extraction process, rather than a vocal based alone. To the best of our knowledge, we are the first to tackle this variant on the SD problem (or CPD) from the intelligent NLP standpoint, and with a dataset in the Hebrew language which is an issue in its own right. We empirically show, based on two distinct datasets, that our method is abled to accurately identify the CPD's in an audio burst with 82.12\% and 89.02\% of success in the Recall and F1-score measurements.
	\end{abstract}
	
	\section {Introduction} \label{introd}
	Alice asks Bob to make a statement. Two days later, she decides to go over his answers because she feels she missed some details. As Bob likes to talk, and since Alice went over the details carefully, the recorded statement lasted more than an hour. Luckily, Alice has access to a diarization system that allows her to isolate the segments in which Bob was talking. A Speaker Diarization (SD) system aims to  answer the question “who spoke when?” for a given audio burst, by identifying the start and ending time of each segment, as well as the speaker's identity. Speaker Diarization~\cite{wan2018generalized, wang2018speaker, zhang2019fully} has many applications in real world: speaker turn analysis, speaker indexing, speech recognition with speaker identification, and diarizing meeting and lectures \cite{pardo2007speaker, reynolds2004lincoln}. Given the scope of the diarization problem (e.g., unknown number of speakers, supporting multiple languages, voice quality), SD is considered as a hard research problem, that has attracted numerous researcher's attention for more than a decade ~\cite{yella2014overlapping, soldi2015adaptive, kenny2010diarization, el2009improved, himawan2018investigating, dimitriadis2017developing, cyrta2017speaker, hruz2017convolutional}.
	\\\\
	A typical SD system can be divided into a number of components: (i) a speech segmentation module, which detects and removes the non-speech parts, then divides the input into small segments; (ii) An embedding extraction module, where speaker-discriminative embeddings such as speaker factors or Mel Frequency Cepstral Coefficient (MFCC) vectors ~\cite{sunitha2015speaker} are extracted from each segment; (iii) a clustering module, which determines the number of speakers, and (iv) a classification module that assigns a speaker identity for each segment. Since our objective is to detect the time points at which the identity of the speaker changes, we actually solve the Change Point Detection problem (CPD) ~\cite{camci2010change, chatterjee2017optimal, hartland2007change}. In statistical analysis, the CPD problem consists of identifying locations on time when the probability distribution of a stochastic process or sequential data changes. More generally, the CPD problem detects whether or not a change has occurred, and identifies the times when these changes took place. One of the main CPD applications is the detection of anomalous behavior, which is well known in literature as the Anomaly Detection (AD) problem ~\cite{kieu2018outlier, intrator2018mdgan}. In a typical anomaly detection system, a clustering algorithm is needed to identify the set of outliers ~\cite{kieu2018outlier}. Since the CPD problem can be easily framed as an AD, this work can be seen as an AD problem as well, because the change points in our data are exactly the anomalies that we are looking for. 
	\\\\This paper presents the benefits of tackling the CPD problem from a textual standpoint. We introduce a new model for the CPD (or AD) problem called \textbf{TSCPD}, which stands for \textbf{T}extual \textbf{S}peaker \textbf{C}hange \textbf{P}oint \textbf{D}etection. We focus on the first task (i.e., division into segments), and empirically show the negative correlation between an increase in the number of speakers and the state-of-the-art solution quality. Thus, we propose a transition from the vocal domain to the textual domain by running a Speech-2-Text algorithm over the dataset on the vocal signals.
	One of the main contributions of the \textbf{TSCPD} approach, is that it yields a robust model, which is only trained by a neural network once over the dataset, but then fits for other datasets with speakers who did not appear in the dataset used to train the model at all (see Section~\ref{unseen_speakers}). In addition, our model shows robustness on other Speech-2-Text engine and new speakers (see Section~\ref{new_s2t_data}). More specifically, we present an NLP-based hybrid solution for the CPD problem, as well as the development process of a deep learning based approach such that given an audio burst, our method determines all of the points in time in which the speaker identity has changed. 
	\\\\
	 To the best of our knowledge, we are the first to propose an intelligent NLP based solution that (I) tackles the CPD problem with a dataset in Hebrew (discussed in depth in Section~\ref{related_work}), and (II) solves the CPD variant of the SD problem. Our empirical experimental evaluation process (Section~\ref{exp_eval}) shows that the transition from voice analysis to the NLP approach, can accurately identify the CPD's for a given multi-participant conversation with 82.12\% success reflected by the model's Recall measurement. Moreover, our speakers-number-independent approach outperforms recently voice-based solutions~\cite{zhang2019fully, wan2018generalized} to the original SD problem, which are based on the D-vectors~\cite{jung2017d, heigold2016end} extracted from the speech signal and clearly depend on the number of speakers. In addition, while the number of speakers and their differences in pronunciation and intonation are a critical issue in these voice-based solutions, we show that our approach remains robust to these vocal elements. 
	\\\\The  remainder  of  this  paper  is  structured  as  follows: Section~\ref{related_work} surveys related work on the famous SD problem, clustering algorithms and the difficulties of coping with datasets in the Hebrew language; Section \ref{data} provides a brief background on the dataset and a description of the feature encoding process; Section \ref{our_approach} presents the methodology employed in this work, and our Neural Network based approach; Section \ref{exp_eval} provides a fine-grained experimental evaluation framework, and Section \ref{conclusions} concludes and  summarizes this paper. For ease of reading, we provide a list of abbreviations in Table~\ref{tab:list_of_abbreviations}.
	\\\\
		\begin{minipage}  {\linewidth} 
		\centering
		
		\begin{tabular}  {C{1.85in} *4{C{3.15in}}}\toprule[1.5pt]
			\bf Abbreviation & \bf Meaning 
			\\\midrule
			SD  &  Speaker Diarization \\
			CPD &  Change Point Detection \\
			NLP &  Natural Language Processing \\
			AD  &  Anomaly Detection \\
			TSCPD  & Textual Speaker Change Point Detection \\
			SV  &  Speaker Verification \\
			SCD & Speaker Change Detection \\ 
			UIS-RNN & Unbounded Interleaved-State Recurrent Neural Networks \\
			\bottomrule[1.25pt]
			\end {tabular}\par
			\bigskip
			\captionof{table}{List of Abbreviations} \label{tab:list_of_abbreviations}
	\end{minipage}
	
	\vspace{\belowdisplayskip}

	\section{Related Work} \label{related_work}
    
    The SD problem and its variants have been discussed in numerous research papers for the last few years  ~\cite{wang2018speaker, zhang2019fully, wan2018generalized, heigold2016end, balamurugansurvey, das2017one}. It is striking that most works on the SD problem have tackled it from the speech signal point of view. The model presented in~\cite{zhang2019fully}, aims to solve the SD problem through a supervised speaker diarization approach, termed Unbounded Interleaved-State Recurrent Neural Networks (UIS-RNN). The UIS-RNN approach is designed to solve the SD problem by learning the extracted speaker-discriminative embeddings (a.k.a. D-vectors~\cite{jung2017d, heigold2016end}) from input utterances, where each individual speaker is modeled by a parameter-sharing RNN, whereas the RNN states for different speakers interleave in the time domain. This RNN is naturally integrated with a distance-dependent Chinese Restaurant Process (ddCRP, ~\cite{blei2011distance, ahmed2008dynamic}) to accommodate an unknown number of speakers. This UIS-RNN model draws on the construction in~\cite{wan2018generalized} of D-vectors, whose authors developed both a Text-Dependent and a Text-Independent Speaker Verification (TD-SV, TI-SV) method for the Speaker Verification (SV) problem, based on a new loss function called Generalized End-to-End (GE2E) loss, which makes the training of speaker verification models more efficient than Tuple-based End-to-End (TE2E,~\cite{heigold2016end}) loss function. In this GE2E method, the training process consists of building training batches from $N$ different speakers and $M$ speech utterances for each of the $N$ speakers at each step, so that each speaker utterance is represented as a feature-vector. Then, the features extracted from each speaker utterance are fed into a Long-Short-Term-Memory (LSTM~\cite{hochreiter1997long}) Neural Network. Next, a linear layer is connected to the last LSTM layer as an additional transformation of the last frame response of the network. The outcome of the training process is an embedding vector (D-vector) for each speaker utterance, which is defined as the L2 normalization of the neural network output. By contrast, the model presented in~\cite{fini2019supervised} suggested using X-vectors for the online SD problem, by performing qualitative modifications of the UIS-RNN~\cite{zhang2019fully} to improve learning efficiency and the overall diarization performance. In particular, they introduced a loss function called Sample Mean Loss (SML), as well as a modelling of speaker turn behavior, which involved devising an analytical expression to compute the probability of a new speaker joining the conversation.
	\\\\
	We show that these state-of-the-art SD systems~\cite{zhang2019fully, wan2018generalized} suffer from a dramatically lower recall as the number of speakers in the dataset increase. Clearly, this dependency poses a massive challenge, especially since the number of speakers in a given audio burst is typically unknown in advance. This incomplete information forces a typical SD system to rely on the performance of clustering tools ~\cite{lin2019lstm, shum2013unsupervised, sell2014speaker, mansfield2018links} to determine the number of speakers in a given audio burst. Assuming that a clustering process correctly detects how many speakers are in a given audio burst, an SD systems quality is assessed in terms of the assignment of speaker identities to each segment in the audio burst. The two other main challenges in SD are language variants, which are manifested in the speaker pronunciation and intonation, and the sound quality. All of these, impact the speech signal analysis. 
	\\\\
	The solutions in~\cite{zhang2019fully, wan2018generalized, heigold2016end} have tackled the SD\textbackslash SV problems in terms of the speech signal. Other examples related to SD and the speech signal are presented in~\cite{laufer2018diarization} that described a SD method based on the separation of under-determined speech mixtures, where the number of speakers was greater than the number of microphones; The authors in~\cite{brendel2019localization} proposed an algorithm to estimate the number of speakers, using reliability information to obtain robust estimation results in adverse acoustic scenarios, and estimating the individual probability distributions describing the location of each speaker using convex geometry tools. However, one component remains unsupervised in most modern SD systems, as well as in the one presented in~\cite{zhang2019fully}; namely, the clustering module such as Gaussian Mixture Models (GMM)~\cite{zajic2017speaker}, k-means~\cite{dimitriadis2017developing}, etc. The main drawback of the dependency on a clustering module, is the likelihood of choosing the wrong number of clusters. In an SD system the situation is more critical, since the more speakers in a given audio burst, the higher the probability of weakening the solution quality and wrongly classifying speaker identities to speech utterances.
	\\\\Another main drawback of clustering algorithms is their sensitivity to outliers, which has been the focus of interest in problems such as the one presented in this work, which is a variant of the CPD~\cite{truong2019selective} problem. The CPD can easily be formulated as a text Speaker Change Detection (SCD) problem, since that the segmentation process in our case (textual context) is considered as an outlier. This can be seen in~\cite{india2017lstm} which presented an SCD system based on LSTM Neural Networks using both acoustic data and linguistic content, or in~\cite{luo2010segmentation} which developed a segmentation based algorithm for text-dependent speaker recognition, for real-time applications on embedded platforms. Additional approaches for the textual SCD were presented in~\cite{zajic2018recurrent}, that designed a textual based solution for the textual SCD by feeding an LSTM Neural Network with a one hot encoding vectors, which is definitely inappropriate to our method (see Section~\ref{our_approach}) as our dataset contains 81,301 different words, and in~\cite{meng2017hierarchical} that formulated the text-based SCD as a matching problem of utterances before and after a certain decision point, then proposed a hierarchical Recurrent Neural Network (RNN) with static sentence-level attention. Moreover, the method presented by~\cite{meng2017hierarchical} can't maintain robustness to new speakers (see Section~\ref{unseen_speakers}).
	\\\\
	The CPD problem has attracted researchers' attention over the last few decades, and in fact the first works on the CPD problem date back to the 1950s~\cite{page1954continuous, page1955test}, in works aiming to locate the shift in the mean of Independent Identically Distributed (IID) Gaussian variables for industrial quality control purposes. The CPD problem has been investigated in terms of (i) Speech Processing~\cite{harchaoui2009regularized, wang2018voicefilter} for the task of audio segmentation; (ii) Financial Analysis~\cite{lavielle2007adaptive}, for adaptive detection of multiple change-points in asset price volatility; (iii)  Bio-Informatics~\cite{vert2010fast} in the approximation of a multidimensional signal by a piecewise-constant signal, using quadratic error criteria, or when confronted with several 1-dimensional signals which probably and reasonably share common in patient's genomic profiles. The attempt to solve this CPD problems from such a wide range of disciplines gave raise to the specific field of AD~\cite{nicolau2016hybrid, chalapathy2019deep}. Even though CPD and AD have certain research objectives in common, there is still a difference between the two, mainly as regards efforts to identify events or observations which raise suspicions because they differ significantly from the majority of the data. This difference is emerged clearly when tackling AD for networking intrusion as presented in~\cite{dokas2002data}, or even in the case of formulation of  abstract methods for many special data types that are usually handled by specialized algorithms~\cite{schubert2014local}.
	\\\\All of these works use SD, CPD, AD in slightly different ways. Since none of them has anything to do mainly with the textual aspect, we dubbed our textual solution termed TSCPD which refers to the speaker segmentation variant of the CPD problem and is based on an intelligent NLP technique.

	\section{Dataset} \label{data}
	The dataset used for this research was composed of 1,692 vocal audio bursts in the Hebrew language, with an a~priori unknown number of speakers in each audio burst. The audio bursts are heterogeneous and include recordings of TV shows, radio programs and TV broadcast, involving 1,240 different speakers. Using a commercial Speech-2-Text engine (publicly available at: \href{https://www.almagu.com/voicetotext}{Almagu-Website}), we converted the audio bursts into 1,692 textual conversations, such that each textual utterance of $n$ words was represented as $n$ rows, where each row contains four columns: the word, the speaker's identity~\footnote{The identity of the speaker was labeled using human experts. Note that this labeling is mainly for comparison with SD systems and our solution only requires a binary classification (i.e, same speaker as before or not) for the training process.}, the start and the end times of the word. A natural obstacle we were faced with while exploring the dataset was class imbalance; A quick statistical analysis showed that 98.5\% of the samples were labeled as non-CPD examples (i.e., rows $i$ and $i+1$ were tagged by the same speaker identity for some arbitrary row $1 \leq i\leq n-1$), whereas only 1.5\% of the  samples were labeled as CPD examples. In Section~\ref{exp_eval} we discuss this challenge in detail and show how to wisely address this obstacle. Another obstacle we were faced with was the dataset language. Although we present a general method which does not assume any a-priori knowledge or rules system for the dataset, there are some difficulties when analyzing a dataset in Hebrew  that do not exist in English or in any Latinate languages; The word order for example in Hebrew is very flexible and not constant. For instance, in Hebrew the verb can appear either before or after the subject of the sentence. Another flexibility that makes Hebrew harder to analyze (e.g. compared to English), are function words such as $"in"$, $"from"$, $"that"$ or $"to"$. In Hebrew, function words are concatenated to the following word instead of being separated with white a space, which thus significantly enlarges the lexicon (vocabulary size) of the Hebrew language in terms of automatic processing. Verbs are another instance of difficulty; In English, the phrase $"when"$ $"they"$ $"came"$ is made up of three words, whereas in Hebrew, one lexeme can reflect the (i) verb, (ii) function word, and (iii) singular and plural relationship. Thus, complex analysis and structural elements such as morphology~\cite{goldberg2013word, more2016data, seddah2013overview} needed to be parsed to understand the meaning of a sentence in Hebrew. In Section \ref{feature_eng} we show how we dealt with the issues arising from a (transformed from speech signal) textual dataset in Hebrew, mainly by tackling its morphological features and ambiguity.
	
	\subsection{Feature Engineering - Data Encoding} \label{feature_eng}
\begin{figure}[!ht]
    	\centering
    	\includegraphics[scale=0.65]{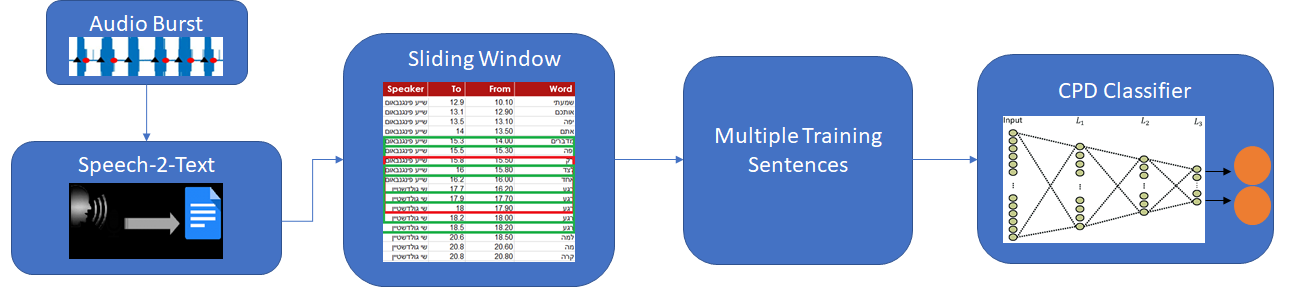}
    	\caption{An illustration of the Sliding-Windows method applied over the transformed textual conversations, which were then fed as training samples into a Fully-Connected Neural Network that classified whether a given instance was a CPD or not.}
    	\label{fig:1}
    \end{figure}

	To transform all textual conversations into a training and test sets, we transformed and iterated over all the conversations using the framework presented in Figure~\ref{fig:1}.           
	\\
	
		\begin{figure}[!ht]
	\centering
    	\begin{subfigure}[b]{0.82\textwidth}
    	    \captionsetup{width=1.3\linewidth}
    		\includegraphics[width=\linewidth]{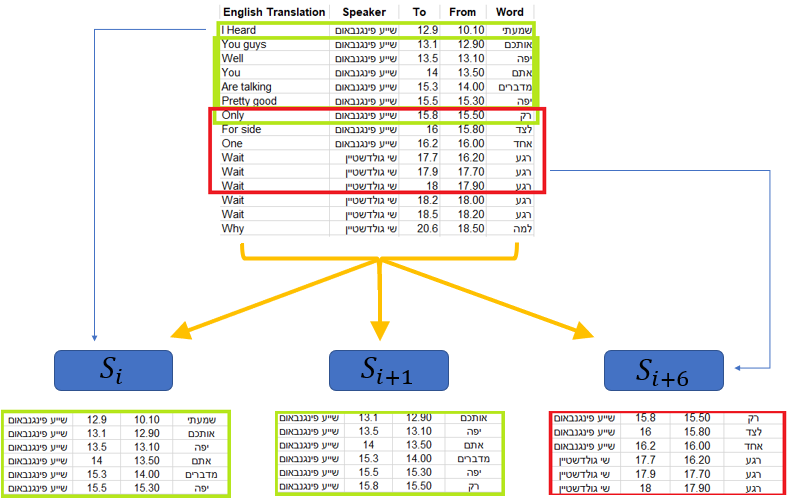}
    		\caption{An illustration of the Sliding-Window method applied over a given textual conversation (after the Speech-2-Text transformation), and iterative computation of Sliding-Windows through the continuous text. The red frame ($S_{i+6}$) represents an example of a speaker interchange; i.e., a change point, whereas the green frames ($S_{i}, S_{i+1}$) are an example of a non change point examples. The column \textbf{English Translation} stands for translation from Hebrew to English (not in the dataset, used solely for the purposes of this paper). The challenges presented in the Related Work (Section~\ref{related_work}) related to Hebrew analysis are listed in this column for each utterance or phrase.}
    		\label{fig:2a}
    	\end{subfigure}
   
    	\begin{subfigure}[b]{1.0\textwidth}
    	    \vspace{10pt}
    	    \captionsetup{width=1.06\linewidth}
    		\includegraphics[width=\linewidth]{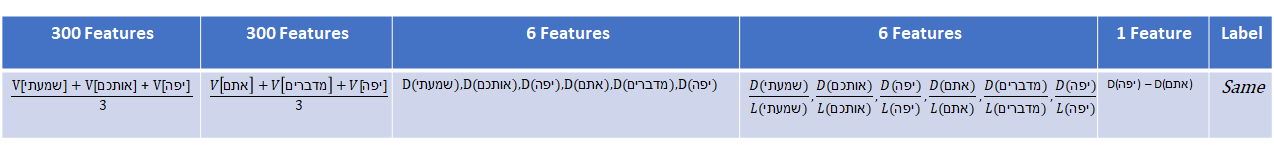}
    		\caption{Computation of a concatenated Feature-Vector $V_i$ for the Sliding-Window $S_i$ presented in Figure~\ref{fig:2a}; $V_i$ was concatenated by - (i) the vectorial average representation of $S_i^I$, (ii) the vectorial average representation of $S_i^{II}$, (iii) the duration of each of the words in $S_i$, i.e. $\{W_i^{j}\}_{j=1}^{6}$, denoted by $D(W_i^j)$, (iv) the ratio of the duration of each word in $S_i$, and its length in characters (denoted by $L(W_i^j)$), and (v) the time elapsed between $W_i^3$ and $W_i^4$ in $S_i$. For illustrative purposes, we added the label of $S_i$ ($Split$).}
    		\label{fig:2b}
    	\end{subfigure}
    		\caption{Creation of training instances using the Sliding-Window method, and illustration of the feature encoding process. Each column represents the number of cells concatenated to the feature vector.}
    \end{figure}

	For each conversation of size $n$, we iterated  over all of the $n$ words (see Figure~\ref{fig:2a}) and divided the conversation into tuples of six words, using the sliding window method. This resulted in $n-5$ tuples (windows), denoted as S=\{$S_1$,$S_2$,\ldots,$S_{n-5}$\}. The first window, $S_1$, was composed of the first six words ($W_1^1, W_1^2,.., W_1^6$). Similarly, $S_2$ was composed of the next six word tuple ($W_2^1, W_2^2,\ldots, W_2^6$) and so on to the last window, $S_{n-5}$ which included ($W_{n-5}^1, W_{n-5}^2, \ldots, W_{n-5}^6$). For each arbitrary sliding window $S_i$ ($1\leq i \leq n-5$), we computed the Word-2-Vec~\footnote{A group of models that are used for producing word embedding, by two-layer neural networks that are trained to reconstruct linguistic contexts of words.}~\cite{mikolov2013distributed, mikolov2013efficient} vectorial average of the first and last three words in the sliding window $S_i$, denoted by $S_i^I=\{W_i^1,W_i^2,W_i^3\}$ and  $S_i^{II}=\{W_i^4,W_i^5,W_i^6\}$, respectively. For the Word-2-Vec based computation, we used the Word-2-Vec pre-trained vectors vocabulary file in Hebrew, published by Facebook (available at \href{https://fasttext.cc/docs/en/crawl-vectors.html}{Facebook pre-trained Word-2-Vec Models}). Since every word in the vocabulary was represented as a multi-dimensional vector of size $300$, this step in the feature engineering process generating 600 features (300 for $S_i^I$, and 300 for $S_i^{II}$. In addition, even though our solution was mainly NLP based, our feature-engineering process generated additional $13$ meta-vocal based features, for each of the six words in $S_i$, as follows: 
	
	\begin{enumerate}
	    \item[1-6.] Duration of each word in $S_i$.
	    
	    \item[7-12.] Speech rate of each word, represented by the length of a given word (in characters), divided by its duration.
	    
	    \item[13.] Time elapsed between the third word $W_i^3$, and the fourth word $W_i^4$ in $S_i$, that can intuitively imply about some percentage of confidence in the detection of change point. 
	\end{enumerate}
	Figure~\ref{fig:2b} visualizes the full feature vector of size 613 engineered for each Sliding-Window.
	Finally, for each window, we added the label for the target feature (i.e., $Same$ or $Split$). If there was a change of speakers between the third and fourth word, this instance was labeled $Split$. Otherwise, the label was $Same$.

	\section{Classification Using a Fully-Connected Neural Network} \label{our_approach}
\begin{figure}[!ht]
    	\centering
    	\includegraphics[scale=0.33]{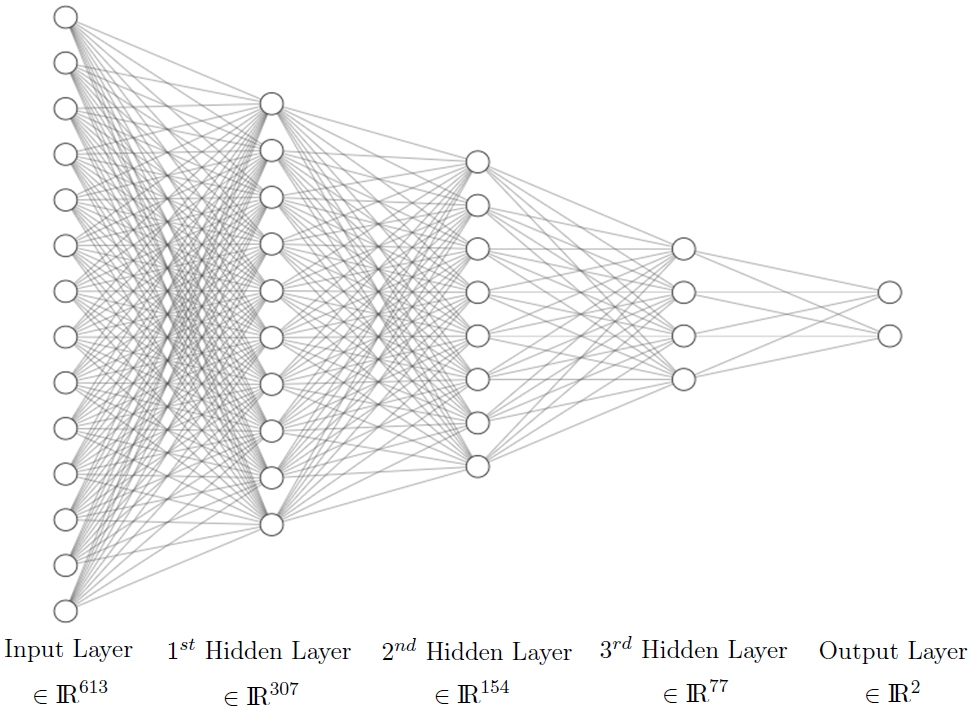}
    	\caption{A visual depiction of the Neural Network with a Fully-Connected architecture and the dimensions of each layer (via a web-tool available at the \href{http://alexlenail.me/NN-SVG/index.html\#}{NN-SVG tool}). The Neural Network has three hidden layers with a dropout layer between each hidden layer and its following layer, as well as between the Input Layer and the $1^{st}$ Hidden Layer; the dropout layers are only reflected during the Neural Network's training process, with the probability of $p=0.5$ to remove a random neuron from the Neural Network.}
    	\label{fig:fc_illustration}
    \end{figure}

    Our feature engineering process resulted in 1,129,570 sliding windows instances, which were converted into feature-vector based learning examples. These examples were divided into training and test sets, using the well known Cross-Validation method ~\cite{stone1974cross, kohavi1995study, arlot2010survey}, such that 80\% of the learning examples were used for the training set, and the remaining 20\% for test. Due to the tabular structure of the transformed dataset using the Sliding-Window method, the model chosen for the classification problem was a Deep Neural Network with a Fully Connected (DNN-FC) architecture. The Neural Network has three hidden layers, and a dropout layer between any two layers, as well as between the last hidden layer and the output layer, with $p=0.5$ for all the dropout layers (see Table~\ref{tab:nn_architecture} for a specific input size and Neural Network layer dimension, and  Figure~\ref{fig:fc_illustration} for an illustration of the Neural Network's architecture).
    \\
    \vspace{\belowdisplayskip}
    
    \begin{minipage}  {\linewidth} 
		\centering
		
		\begin{tabular}  { C{1.25in} C{1.85in} *4{C{2.15in}}}\toprule[1.5pt]
			\bf Layer & \bf In Features & \bf Out Features 
			\\\midrule
			Input             	   &  613     & 307\\
			$1^{st}$ Hidden Layer  &  307     & 154\\
			$2^{nd}$ Hidden Layer  &  154     & 77\\
			$3^{rd}$ Hidden Layer  &  77      & 2\\
			Output             	   &  2       & Softmax Decision Function\\
			\bottomrule[1.25pt]
			\end {tabular}\par
			\bigskip
			\captionof{table}{Neural Network Architecture} \label{tab:nn_architecture}
	\end{minipage}

    Since our problem is an instance of a supervised classification, we used the Cross-Entropy (CE) loss function ~\cite{zhang2018generalized}. This loss function measures the performance of a classification model whose output is a probability value between 0 to 1 for each label (i.e. the CE loss increases as the predicted probability diverges from the actual label). As for the learning rate of the Neural Network, we used the Adam Optimizer ~\cite{kingma2014adam, zhang2017normalized}, an adaptive learning rate method that computes individual learning rates for different parameters, with an initial learning rate value of $\alpha=1\cdot 10^{-4}$. Consequently, we chose a Softmax~\cite{qin2019rethinking, qin2019softmax} activation layer to represent the network's output, in the form of a probability vector of size 2 for the Softmax layer which predicts the probability of an instance being labeled as  $Split$ and $Same$, with one probability for each class.

	\section{Experimental Evaluation}\label{exp_eval}
	In this section we validate our hypothesis that the classification framework suggested in this paper results in accurate and effective speaker change point detection. Specifically, we show that using NLP techniques for CPD in a multi-speaker environment can outperform classical speech techniques. For this purpose, we used human experts to tag the dataset by assigning a label of speaker identity to each word, in a given audio burst that was converted into a list of words (with the additional features as the start and end times of the word). Clearly, there is no need in a human labeling process for future dataset classification (apart from evaluation purposes), since the TSCPD model aims to detect the change points, rather than speaker identities of each vocal segment. Unlike typical voice-based solutions (\cite{zhang2019fully, wan2018generalized}) which require the speaker identities, the TSCPD model need to know whether an interchange has occurred or not, independent of the speaker identities before and after the interchange. 
	\\\\The last thing that had to be considered while training the neural network was the ratio of the examples from $Split$ class to the examples from $Same$ class. One could argue that the construction of a classifier which always returns $Same$ as a label would be good, since, this classifier would produce $\sim$ 98\% of precision on our dataset. This is due to the fact that in a given natural conversation, the proportion of speech is higher than the number of turns. Thus, this classifier is of limited utility for our purposes to find and identify change points. To do so, our method takes the relative weight of the examples from class $Split$ into account. As mentioned in Section~\ref{data}, the comparative proportions of the $Split$ and $Same$ classes were 1.5\% and 98.5\% respectively. Thus, when defining the loss function of the neural network, we defined the weight of each class to be the inverse of the number of examples from this class, i.e. $\frac{1}{|Split|}$ for class $Split$, and $\frac{1}{|Same|}$ for class $Same$ (where $|Split|$ and $|Same|$ represent the number of examples from each class). 
	
	The remainder of this section presents the evaluation process as follows:
	\begin{itemize}
	    
	    \item Section~\ref{exp_eval_amount} analyze the robustness of TSCPD model to a variable number of speakers, compared to the voice based diarization solutions presented in~\cite{zhang2019fully, wan2018generalized}.
	    
	    \item Section~\ref{unseen_speakers} analyze the robustness of TSCPD model to a completely new speakers that were not seen during the training process.
	    
	    \item Section~\ref{dl_ml_models} compares the TSCPD model to a Deep Learning approach - an Auto Encoder, and Machine Learning algorithms, such as XG-Boost, SVM, Decision-Tree and k-NN.
	    
	    \item Section~\ref{new_s2t_data} analyze the robustness of TSCPD model to a different source of Speech-2-Text engine.
 	    
	\end{itemize}

        \subsection{The Number of Speakers Influences Model Efficiency} \label{exp_eval_amount}
        The objective of this work was to tackle the CPD problem mainly in terms of its dependence on the number of speakers, which is a key issue when attempting to solve the SD\textbackslash SV problems with voice based solutions. The work presented in~\cite{zhang2019fully} utilized the speech embedding extraction module in~\cite{wan2018generalized}, by learning the extracted speaker-discriminative embeddings, or D-vectors, from input utterances. In order to show our model's robustness to the number of speakers, and compare our work to the speech embedding extraction module, we translated the SD problem into a CPD. For this purpose, we compared our model and the one presented in~\cite{zhang2019fully}, with a variable number of speakers $k\in\{8, 20, 50, 100, 200, 250, 500, 1000\}$, and with 10 different random speakers sampling's identities and utterances, using the following procedure:
		
		\begin{enumerate}
		    
		    \item Step one consisted of splitting our dataset into a textual speaker utterances. Since each word duration is known (by substracting the "to" from the "from" columns), as well as the speaker identity, we could retrieve a textual speaker utterance for each one in the dataset. 
		    
		    \item Step 2 consisted of segmenting each audio file according to its timeline speaker utterances. Then, we trained the neural network presented in~\cite{wan2018generalized}, to extract the speech embedding vectors for each speaker utterance, i.e. the D-vectors. 
		    
		    \item The D-vectors extracted for each speaker utterance formed a matrix of size $N_{seg} \times M_{emb}$, where $N_{seg}$ represents the number of segments extracted for each speaker utterance, and $M_{emb}$ is the D-vector which $\in \rm I\!R^{256}$. The segments for each speaker utterance were extracted using a Voice Activity Detector (VAD) engine, using the $webrtcvad$ and $librosa$ Python libraries.
		    
		    \item Then, we split our converted to speech utterances dataset; i.e., the speech embedding matrices for each speaker utterance, into training and test sets with a 80\%-20\% division of the data samples respectively. Clearly, the number of data samples for each speaker identity out of the total of 1,240 was assigned proportionally to the training and test sets.  
		    
		    \item The work in~\cite{zhang2019fully} presents a fully-supervised speaker diarization model termed UIS-RNN, and assigns a speaker identity to each speaker utterance in the dataset. As such, we trained the UIS-RNN presented in~\cite{zhang2019fully}, by the training matrices, each of which represented a set of D-vectors extracted as function of the number of segments in each speaker utterance.
		    
		\end{enumerate}
		
		For each number of speakers $k\in\{8, 20, 50, 100, 200, 250, 500, 1000\}$, we ran this procedure 10 times, each time with a randomly chosen set of speakers. In order to calculate the error rate of each result for each number of $k$, we converted the model's inference results into CPD based results, i.e. a binary classification testing. For each speaker utterance in the test set, the UIS-RNN model produced a classification vector of size $N_{seg}$. Hence, we could conduct \textbf{Type I}, \textbf{Type II} ~\cite{banerjee2009hypothesis} errors analyses for each classification vector, which later resulted in Precision, Recall and F1-Score calculations. For a given conversation split into $U\in\mathbb{N}$ speaker utterances, the \textbf{Type I} and \textbf{Type II} errors analysis was as follows:
		
		\begin{itemize}
		
		    \item \textbf{Type I}  -  occurs any time when in a given speaker utterance $u_i$ (for $1\leq i\leq |U|$), the first segment label in its corresponding classification vector had different values than each of the classification vector elements. More specifically, for a classification vector $c_i$ (that represents the identity assignments for $u_i$ segments) of size $\ell$, we count an error each time $c_i[j]$ is not equal to $c_i[1]$ for $2\leq j \leq \ell$. Each such mismatch implies that the UIS-RNN detected a change point and tagged $"Split"$, rather than $Same$.
		    
		    \item \textbf{Type II}  - occurs any time when the last segment of a speaker utterance $u_i$ classification, and the first segment of $u_{i+1}$ classification are equal (for $1\leq i\leq |U|-1$). This is due to the fact that between any proceedings $u_i$ and $u_{i+1}$ there must have been an interchange, i.e. the UIS-RNN tagged $Same$, rather than $Split$. 
		    
		\end{itemize}
		
		\begin{table*}[!ht]
			\centering
			\begin{tabular}{ |p{1.8cm}|p{7.9cm}|p{7.3cm}|  }
				
				\hline
				\multicolumn{3}{|c|}{Random Model List - \textbf{TSCPD, UIS-RNN}} 
				\\
				\hline
				\textbf{Speakers} & {\textbf{TSCPD}} \newline Precision \quad \quad \quad Recall \quad \quad \quad\quad F1-Score  
				&\textbf{UIS-RNN} \newline Precision \quad \quad Recall \quad \quad \quad \quad F1-Score   \\
				\hline\hline
				\centering{\textbf{8}} & 96.56(1.03)\quad\quad 95.11(0.94)\quad\quad 95.78(0.85) & 95.94(1.11)\quad 97.74(0.93)\quad \quad 96.65(1.36) \\
				
				\centering{\textbf{20}} & 96.34(1.87)\quad\quad 94.43(3.36)\quad\quad 95.30(2.55) & 94.87(2.53)\quad 96.67(1.82)\quad\quad  95.15(2.75) \\
				
				\centering{\textbf{50}} & 96.54(0.55)\quad\quad 91.21(2.34)\quad\quad 93.55(1.51) & 94.56(1.75)\quad 96.60(0.92)\quad\quad 94.98(1.34) \\
				
				\centering{\textbf{100}} & 96.90(0.45)\quad \quad 87.42(1.50)\quad\quad 91.50(0.88) & 94.26(1.17)\quad 95.83(0.63)\quad\quad 95.03(1.00) \\
				
				\centering{\textbf{200}} & 97.09(0.22)\quad \quad 84.50(2.22)\quad \quad 89.79(1.41) & 94.48(1.42)\quad 92.14(0.87)\quad\quad 93.29(1.32) \\
				
				\centering{\textbf{250}} & \textbf{97.03(0.20)}\quad \textbf{83.72(1.16)}\quad \textbf{89.30(0.79)} & 95.84(0.32)\quad 82.05(0.55)\quad\quad 87.76(0.25) \\
				
				\centering{\textbf{500}} & \textbf{97.17(0.13)}\quad \textbf{83.44(1.56)}\quad \textbf{89.15(0.94)} & 96.05(0.62)\quad 79.40(1.13)\quad\quad 86.18(0.63)  \\
				
				\centering{\textbf{1000}} & \textbf{97.13(0.16)}\quad \textbf{82.77(0.95)}\quad \textbf{88.71(0.60)} & 96.04(0.39)\quad 79.21(1.16)\quad\quad 86.04(0.70) \\
				
				\centering{\textbf{1240}} & \textbf{97.19(0.00)}\quad \textbf{82.12(0.00)}\quad \textbf{89.02(0.00)} & 95.72(0.00)\quad 79.37(0.00)\quad\quad86.78(0.00)\\
				
				\hline
			\end{tabular}
			
			\caption{Results table for the randomly chosen speaker process, a comparison of our model versus Google's D-vector based model. For each number of speakers $k\in\{8, 20, 50, 100, 200, 250, 500, 1000\}$ the average values of the Precision, Recall and F1-Score are presented for 10 different data samplings. For 1,240 speakers we ran the process only once, for purposes of a full comparison (TSCPD and UIS-RNN models). The Standard-Deviation $\sigma$ appears in parentheses.} 
			\label{tab:random_speakers_table_comparison}
		\end{table*}

		\begin{figure}[!ht]
        	\centering
        	\includegraphics[scale=0.45]{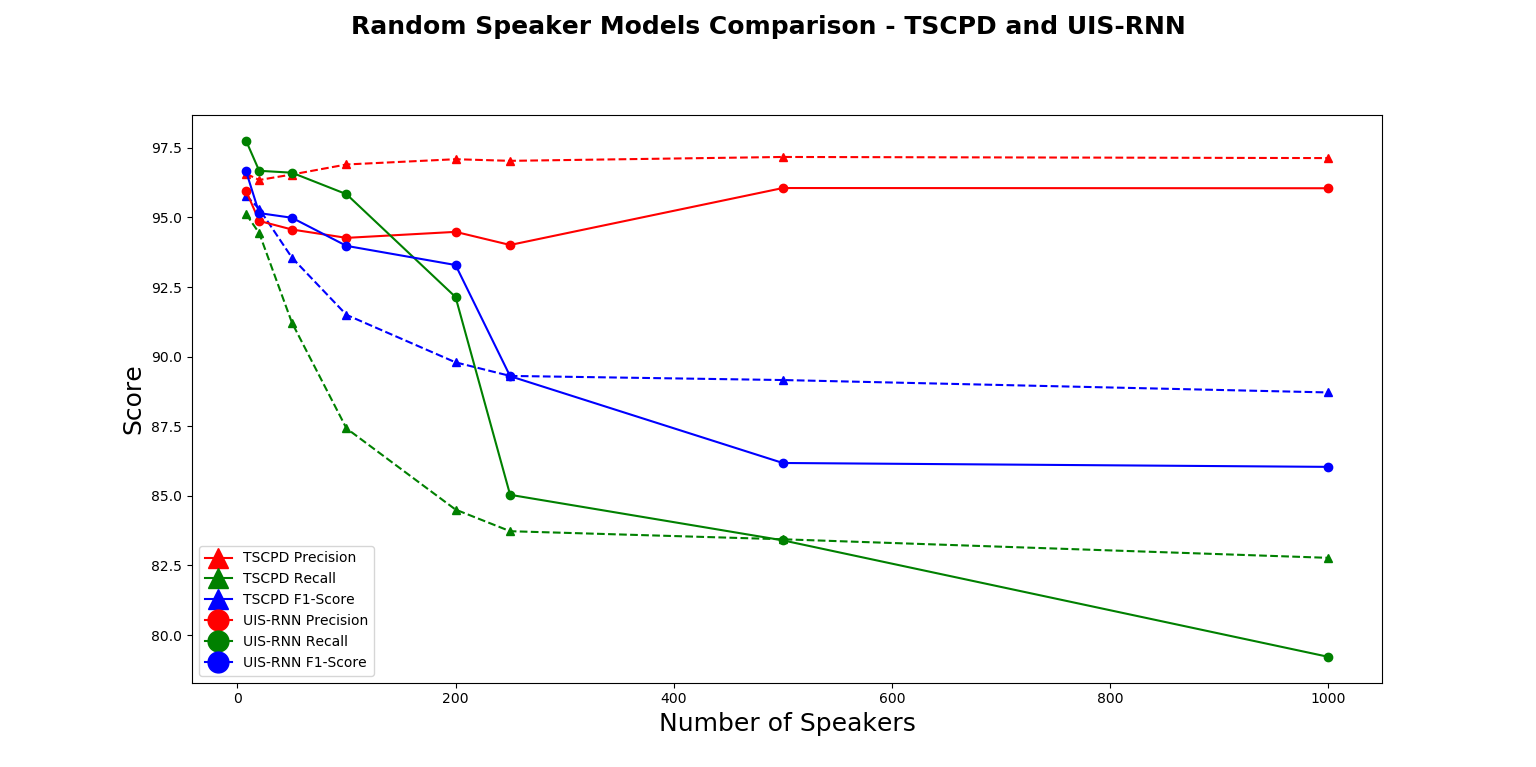}
        	\caption{Graphical results for the comparison process performed as explained in Sub-Section~\ref{exp_eval_amount}. As shown, the Recall and F1-Score drop dramatically when using the UIS-RNN model, whenever that there is an increase in the number of speakers in the dataset.}
        	\label{fig:random_speaker_results}
        \end{figure}

		 The results for this comparison are presented in Table~\ref{tab:random_speakers_table_comparison}, and a graphical illustration of the performance for the TSCPD and UIS-RNN models are presented in Figure~\ref{fig:random_speaker_results}. In addition, as can be seen in Table~\ref{tab:random_speakers_table_comparison}, we compared the models with the whole dataset, i.e. with all of the 1,240 speakers. For this purpose, we ran the described above procedure only once randomly. It is clear that from 250 speakers and up, the TSCPD model outperform the UIS-RNN model in $all$ measurements. Both of the models maintained their Precision due to the extreme class imbalance, but as the number of speakers in the dataset increased - the TSCPD model outperformed the UIS-RNN. The UIS-RNN model's Recall (and as such the F1-Score) deteriorated, since there were many more speaker identities to classify,  whereas the TSCPD remained quite robust to the increase in number of speakers in the dataset. 
		 \\\\
		 The Precision and Recall measures (and as such  the F1-Score\footnote{The F1-Score is defined by the following formula - $F_1 = 2\cdot\frac{Precision\cdot Recall}{Precision+Recall}$}) and Type I \& II errors are very closely related since they are both defined by terms such as true/false positive, and true/false negative. Due to the intuitive and clear numerical representation of the Precision, Recall and F1-Score measurements, we have chosen to report the results by this terms, rather than Type I and Type II errors.
		 \\\\ Figure~\ref{fig:confusion_matrix} presents the Confusion Matrix as well as a graphical illustration of the Receiver Operating Characteristic (ROC). Its corresponding Area Under Curve (AUC) appears in Figure~\ref{fig:roc_auc_curves} for the performance of the TSCPD model for the 1,240 speaker based dataset. The graphs presented in Figure~\ref{fig:roc_auc_curves}, suggest the following: 
		 
		 \begin{itemize}
		 
		     \item As the False Positive value declines, the better the model predicts examples of the $Same$ class with high probability, and vice versa for the True Positive value that correspond to the $Split$ class.
		     
		     \item The TSCPD model is quite robust despite the class imbalance, a fact which is reflected in the superiority of the micro-average ROC curve as well as its area under curve as compared to the macro-average ROC curve (and its area under curve). The robustness is manifested in this superiority, since the macro-average independently computes the metric for each class, and then takes the average result (hence treating all classes equally), whereas a micro-average aggregates the contributions of all classes to compute the average metric.
		 	 
		 \begin{figure}[!ht]
        	\centering
        	\includegraphics[scale=0.95]{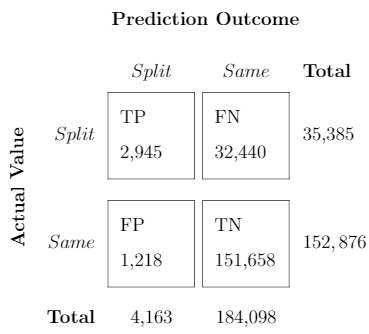}
        	\caption{Confusion Matrix results achieved by the TSCPD model for 1,240 speakers. We treat the $Split$ as the positive class, and vice versa for $Same$. Note the high precision for both classes, even though the positive class barely represents 1.5\% of the dataset examples.}
        	\label{fig:confusion_matrix}
        \end{figure}

		\begin{figure}[!ht]
        	\centering
        	\includegraphics[scale=0.85]{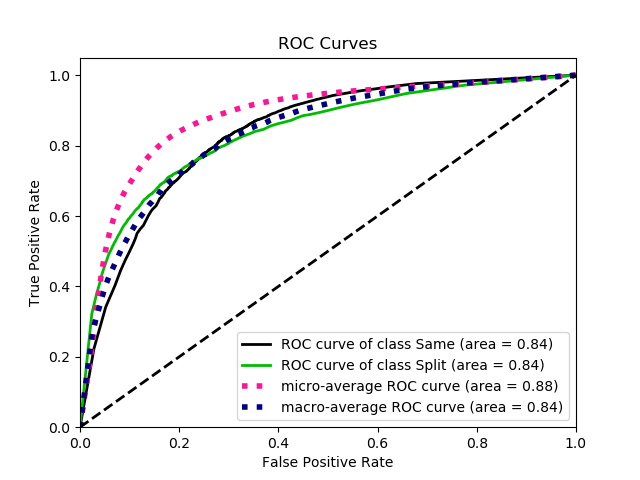}
        	\caption{ROC curves for the TSCPD model performance for 1,240 speakers, as well as the AUC for each curve (see Legend). The black and green curves represent the True Positive Rate as a function of the False Positive Rate for the $Same$ and $Split$ classes respectively. The dashed blue and pink curves represent the same relationship for the computation of micro and macro averages.}
        	\label{fig:roc_auc_curves}
        \end{figure}
		 \end{itemize}

		\subsection{Model Robustness to Unseen Speakers} \label{unseen_speakers}
\begin{figure}[!ht]
        	\centering
        	\includegraphics[scale=0.55]{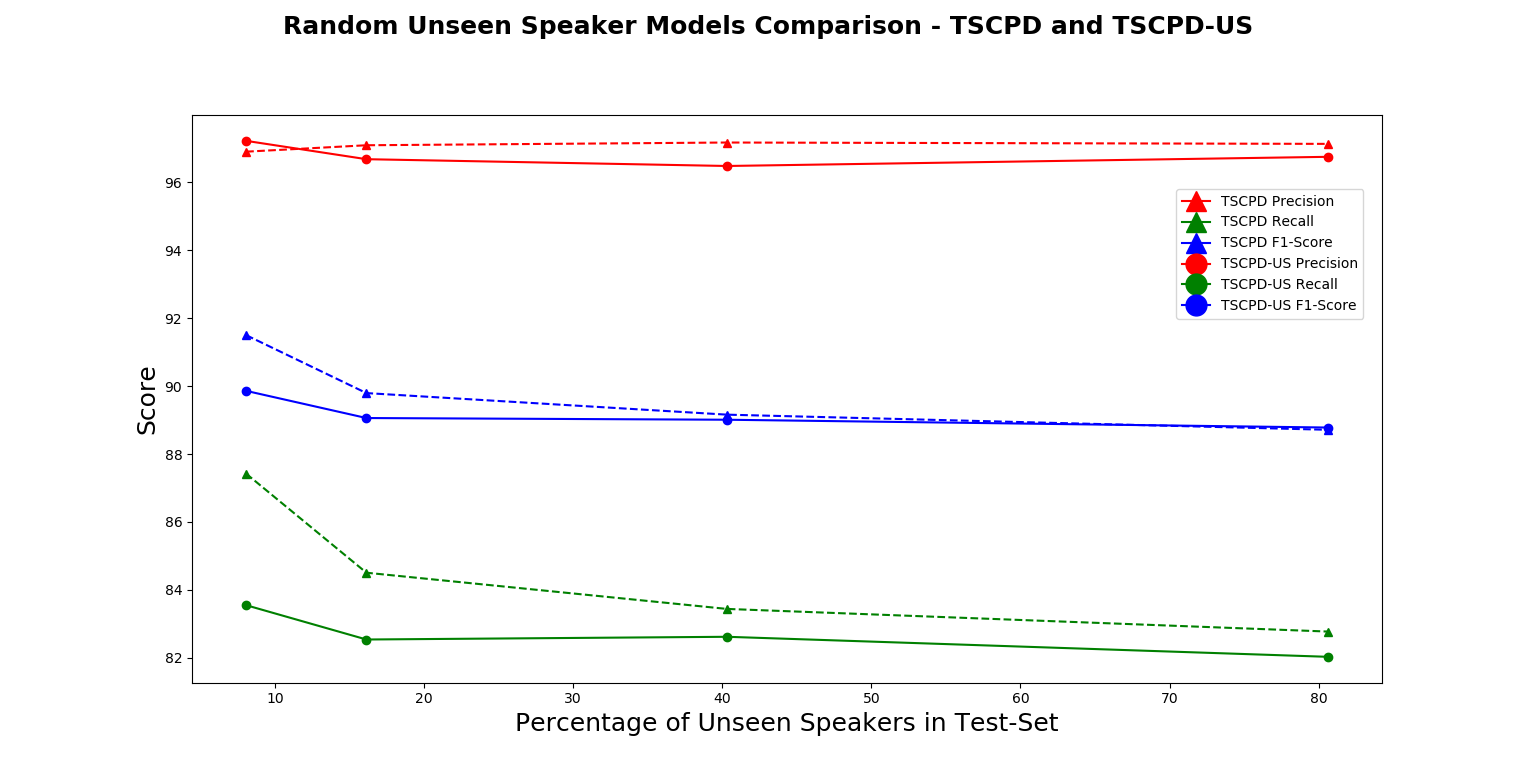}
        	\caption{Graphical illustration of the comparison the TSCPD and TSCPD-US models, as described in Sub-Section~\ref{unseen_speakers}. The X-axis represents the percentage of speakers on whom the TSCPD model was not trained on, i.e. the percentage of new speakers that were introduced to the system. The Y-axis represents the score for Precision, Recall, and the F1-Score. Note the robustness of the TSCPD and TSCPD-US in the plot.}
        	\label{fig:random_unseen_speaker_figure}
        \end{figure}

		Next, we show the robustness and strength of our context based TSCPD model to speakers that were not included in the train set. This means testing whether the TSCPD model is robust to new speakers with (potentially) new different speech rate, that were not used in training the Neural Network. We call this model \textbf{T}extual \textbf{S}peaker \textbf{C}hange \textbf{P}oint \textbf{D}etection - for \textbf{U}nseen \textbf{S}peakers (TSCPD-US). For this purpose, we have split our dataset into training and test sets, by randomly accumulating CSV files (representing transformed Speech-2-Text conversations) up to $k\in\{100, 200, 500, 1000\}$ unseen-during-training-time speakers for the test set, and the rest of the CSV files containing $1,240-k$ speakers for the training set. This ensured that the Neural Network would be only trained on a subset of speakers that were not part of the conversations that served for the test set. Similar to the procedure described in Section~\ref{exp_eval_amount}, for each value of $k$ we ran this dataset splitting 10 times, each time with randomly sampled speakers. The calculation of the average score for each of the 10 random samplings (for each $k$ value), we showed that performance was comparable (with respect to the TSCPD model) when inserting new unseen speakers in the test process. The results for this section are presented in Table~\ref{tab:unseen_random_speakers}, which depicts the slight degradation whenever we insert new unseen speakers to the CPD system. This is graphically illustrated in Figure~\ref{fig:random_unseen_speaker_figure}. These results suggest that whenever $k\geq 200$ completely new unseen tested speakers, the TSCPD converges to robustness around the F1-Score.
		
		\begin{table*}[!ht]
			\centering
			\begin{tabular}{ |p{1.8cm}|p{7.3cm}|p{7.3cm}|  }
				
				\hline
				\multicolumn{3}{|c|}{Random Models List - \textbf{TSCPD, TSCPD-US}} \\
				\hline
				\textbf{Speakers in Test} & {\textbf{TSCPD}} \newline Precision \quad \quad Recall \quad \quad  F1-Score  
				&\textbf{TSCPD-US} \newline Precision \quad \quad Recall \quad \quad F1-Score   \\
				\hline\hline
				
				\centering{\textbf{100}} & 96.90(0.45)\quad 87.42(1.50)\quad 91.50(0.88) & 97.22(0.23)\quad 83.55(1.22)\quad 89.86(0.63) \\
				
				\centering{\textbf{200}} & 97.09(0.22)\quad 84.50(2.22)\quad 89.79(1.41) & 96.68(0.52)\quad 82.54(0.71)\quad 89.06(0.47) \\
				
				\centering{\textbf{500}} & 97.17(0.13)\quad 83.44(1.56)\quad 89.15(0.94) & 96.48(1.02)\quad 82.62(1.31)\quad 89.01(1.15)   \\
				
				\centering{\textbf{1000}} & 97.13(0.16)\quad 82.77(0.95)\quad 88.71(0.60) & 96.75(0.49)\quad 82.03(0.53)\quad 88.78(0.31) \\
				\hline
			\end{tabular}
			
			\caption{Results for the randomly chosen speakers process, comparing the TSCPD and TSCPD models, that tested on a random unseen number of speakers $k\in\{100, 200, 500, 1000\}$. The Standard-Deviation $\sigma$ appears in parentheses.} 
			\label{tab:unseen_random_speakers}
		\end{table*}
		
		\subsection{Classical Machine and Deep Learning Models} \label{dl_ml_models}
		
		Here, we present the comparison to several Deep and Machine Learning models. Specifically, we used the well known Auto Encoder model~\cite{berthelot2018understanding, li2018learning}, since Auto Encoders tend to be applied for Anomaly Detection problems (like the CPD). Using Auto Encoders to detect anomalies usually involves two main steps:
		\begin{enumerate}
		    
		    \item First, we feed our dataset into an Auto Encoder and tune it until it is well trained to reconstruct the expected output with minimum error. An Auto Encoder is well-trained if the reconstructed output is sufficiently close to the input and if the Auto Encoder is able to successfully reconstruct most of the data this way.
		    
		    \item Then, we feed all our dataset again to our trained Auto Encoder and measure the error term of each reconstructed data point. In other words, we measure how far the reconstructed data point is from the actual data-point. A well-trained Auto Encoder essentially learns how to reconstruct an input that follows a certain format, so if we provide a badly formatted data point to a well-trained Auto Encoder, we are likely to get something quite different from our input and a large error term.
		\end{enumerate}
        
\begin{figure}[!ht]
			\centering
			\includegraphics[scale=0.45]{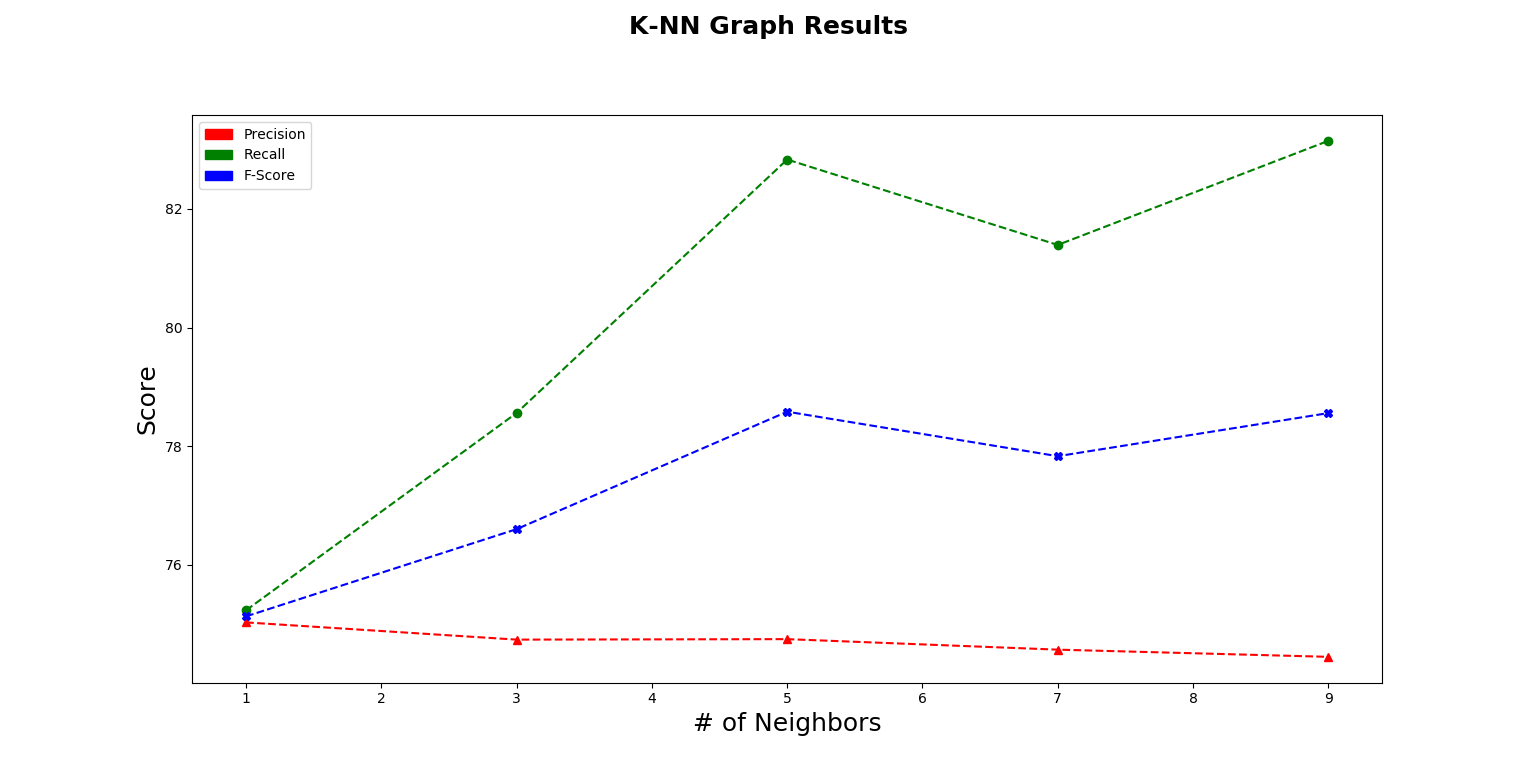}
			\caption{Precision, Recall and F1-Score for  
				 $k$-NN model over the dataset, when $k\in \{1,3,5,7,9\}$. As reported in sections ~\ref{exp_eval_amount}, \ref{unseen_speakers} and \ref{dl_ml_models}, the TSCPD achieved 97.19\% for Precision, 82.12\% for Recall, and 89.02\% for the F1-Score.}
			\label{fig:knn_graph_results}
		\end{figure}

        The Encoder and Decoder parts of the Auto Encoder that we built for this comparison were based on the architecture we used in our Fully-Connected Neural Network, without the $3^{rd}$ hidden and output layers (see Table~\ref{tab:nn_architecture}). In other words, the Encoder was reduced from an input layer size to a $2^{nd}$ hidden layer size, then expanded into the Decoder through the context vector, i.e. from the $2^{nd}$ hidden layer size to the input layer size. However, the Auto Encoder model failed to outperform the TSCPD model.
        \\
        In addition, we examined well-known Machine Learning algorithms to determine whether we could achieve better results than the ones reported in the previous sub-sections. We examined the XGBoost, Support Vector Machine (SVM), Decision-Tree, and $k-NN$ where $k \in \{1, 3, 5, 7, 9\}$. As can be seen in Table~\ref{tab:classical_ml_dl_methods}, neither Auto-Encoder approach, nor traditional Machine Learning algorithms outperformed the TSCPD model. 
        \\\\
        For the different values of $k$ for the k-NN models, Figure~\ref{fig:knn_graph_results} presents the $k-NN$ results graph that sums up the Precision, Recall and F1-Score achieved by using the $k-NN$ algorithm. It is easy to see that the TSCPD model outperformed each of the classical models in terms of Precision and F1-Score, but only achieved lower Recall values in some of the cases (SVM, 5-NN, 9-NN).
        \\
		\begin{table*}[!ht]
			\centering
			\begin{tabular}{|P{5cm}|P{3.5cm}|P{3.5cm}|P{3.5cm}|}
				
				\hline
				\multicolumn{4}{|c|}{\textbf{Machine \& Deep Learning Models Results Table}} \\
				\hline
				\textbf{Model / Indicator} & 
				\textbf{Precision} & 
				\textbf{Recall} &
				\textbf{F1-Score}
				\\
				\hline\hline
				\textbf{TSCPD}	& 97.19 \% & 82.12 \% & 89.02 \%
				\\
				\hline
				\textbf{Auto Encoder}	& 96.81 \% & 78.16 \% & 85.85 \%
				\\
				\hline
				\textbf{XGBoost}	& 74.27 \% & 81.28 \% & 77.62 \%
				\\
				\hline
				\textbf{SVM}	& 74.04 \% & 83.26 \% & 78.38 \%
				\\
				\hline
				\textbf{Decision-Tree}	& 75.18 \% & 74.80 \% & 74.99 \%
				\\
				\hline
				\textbf{1-NN}	& 75.03 \% & 75.23 \% & 75.13 \%
				\\
				\hline
				\textbf{3-NN}	& 74.74 \% & 78.56 \% & 76.60 \%
				\\
				\hline
				\textbf{5-NN}	& 74.75 \% & 82.83 \% & 78.58 \%
				\\
				\hline
				\textbf{7-NN}	& 74.57 \% & 81.39 \% & 77.83 \%
				\\
				\hline
				\textbf{9-NN}	& 74.45 \% & 83.14 \% &  78.55 \%
				\\
				\hline
				
			\end{tabular}
			\caption{Results of the comparison between the TSCPD model and the classical Deep and Machine Learning approaches - the Auto Encoder model for the Deep Learning approach, and XG-Boost, SVM, Decision-Tree and $k$-NN for the Machine Learning algorithms.} 
			\label{tab:classical_ml_dl_methods}
		\end{table*}

		\subsection{New Speech-2-Text Data} \label{new_s2t_data} 
		In Section~\ref{data}, we described the 1,692 vocal conversations that went through the transformation into the text process using a commercial Speech-2-Text engine. In order to further confirm the robustness of the TSCPD model to a new Speech-2-Text engine \footnote{See \href{https://cloud.google.com/speech-to-text}{Google-Speech-2-Text-API-Website}}, we collected an additional 362 vocal conversations with 431 new speakers from similar TV shows and radio programs in Israel, as in our original dataset. These 362 conversations with 431 new speakers were converted using Google’s corpora Speech-2-Text engine, and then transformed into a continuous text, in the same manner as for our initial dataset. Then, we trained and tested the TSCPD model the same way as we did for our initial dataset. 
		\\\\
		The comparison between our dataset and the new Speech-2-Text dataset is presented in Table~\ref{tab:google_s2t_new_data}. It shows that the disparity between the TSCPD model and its performance over the new Speech-2-Text engine manifested mainly in the Recall measurement, and as such over the F1-Score as well. Even though the TSCPD model performance was degraded, it still demonstrated its robustness to a new Speech-2-Text engine.

		\begin{table*}[!ht]
			\centering
			\begin{tabular}{|P{5cm}|P{3.5cm}|P{3.5cm}|P{3.5cm}|}
				
				\hline
				\multicolumn{4}{|c|}{\textbf{Results for the Google Speech-2-Text Data}} \\
				\hline
				\textbf{Model / Indicator} & 
				\textbf{Precision} & 
				\textbf{Recall} &
				\textbf{F1-Score}
				\\
				\hline\hline
				\textbf{TSCPD}	& 97.19 \% & 82.12 \% & 89.02 \%
				\\
				\hline
				\textbf{TSCPD for Google Speech-2-Text Engine}	& 96.82 \% & 76.47 \% & 85.45 \%
				\\
				\hline
			\end{tabular}
			\caption{Comparison of the results for the TSCPD model trained on 1,240 speakers, to the results for the TSCPD model applied to a new dataset with 431 different speakers, transformed using the Google Speech-2-Text engine.} 
			\label{tab:google_s2t_new_data}
		\end{table*}

		\section{Conclusions and Future Work} \label{conclusions}
		In this paper we demonstrated how to solve the CPD variant of the SD problem, and focused on the number of speakers and the conversational context using an intelligent NLP based technique. We showed that we can achieve better results and greater robustness compared to two recently developed voice based solutions to the SD\textbackslash SV problems, and found that our model outperforms other classical Machine and Deep Learning approaches. Even though obtaining better results, the SD problem is still not completely solved. Hence, one possible future research direction would be to address the SD problem with NLP techniques; i.e., to move from the CPD problem towards the full SD problem, including an NLP based assignment to the speaker utterances module. Another possible future work direction would involve addressing the CPD or the SD problems, multi-lingually by solving each problem for many languages by training a cross-lingual model.
		
		\section{Acknowledgment} 
		This work was supported by the Ariel Cyber Innovation Center in conjunction with the Israel National Cyber directorate in the Prime Minister's Office. In addition, we wish to thank to IFAT Group that provided the dataset for this research, as well as the human labeling process for the speaker identities, and express our gratitude to the Israel Innovation Authority (IIA) for funding this study.

		\bibliographystyle{abbrv}  
        \bibliography{main}

	\end{document}